\pdfoutput=1

\documentclass[11pt]{article}

\usepackage{acl}

\usepackage{times}
\usepackage{latexsym}

\usepackage[T1]{fontenc}

\usepackage[utf8]{inputenc}

\usepackage{microtype}

\usepackage{amsmath,amssymb,amsfonts}
\usepackage{algorithmic}
\usepackage{graphicx}
\usepackage{textcomp}
\usepackage{hyperref}
\usepackage{xcolor}
\usepackage{amsmath}
\usepackage{amssymb}
\usepackage{booktabs}
\usepackage{lipsum}
\usepackage{stfloats}
\usepackage{wrapfig}
\usepackage{multirow}

\title{Every picture tells a story: \\ Image-grounded controllable stylistic story generation}

\author{Holy Lovenia\thanks{$^{*}$The authors contributed equally to this work.}\hspace{1.5mm}, Bryan Wilie$^{*}$, Romain Barraud$^{*}$, \\ \textbf{Samuel Cahyawijaya, Willy Chung, Pascale Fung} \\
  Center for Artificial Intelligence Research (CAiRE) \\
  The Hong Kong University of Science and Technology \\
  \texttt{(hlovenia, bwilie, rmbarraud)@connect.ust.hk}}

\begin{document}
\maketitle
\begin{abstract}
Generating a short story out of an image is arduous. Unlike image captioning, story generation from an image poses multiple challenges: preserving the story coherence, appropriately assessing the quality of the story, steering the generated story into a certain style, and addressing the scarcity of image-story pair reference datasets limiting supervision during training. In this work, we introduce Plug-and-Play Story Teller (PPST) and improve image-to-story generation by: 1) alleviating the data scarcity problem by incorporating large pre-trained models, namely CLIP and GPT-2, to facilitate a fluent image-to-text generation with minimal supervision, and 2) enabling a more style-relevant generation by incorporating stylistic adapters to control the story generation. We conduct image-to-story generation experiments with non-styled, romance-styled, and action-styled PPST approaches and compare our generated stories with those of previous work over three aspects, i.e., story coherence, image-story relevance, and style fitness, using both automatic and human evaluation. The results show that PPST improves story coherence and has better image-story relevance, but has yet to be adequately stylistic.
\end{abstract}

\section{Introduction}

Enabling machine-generated stories based on visual cues opens up promising directions, and leads language models (LMs) to be viewed as an interface, allowing its involvement in artistic tasks such as advertisement creation and AI-generated movie scripting~\cite{mcintyre2009learning, ji2022vscript, xu2019clickbait, hao2022language}. 

In that direction, vision-language understanding and generation works succeed in leveraging image as well as text as cross-modal knowledge to solve various tasks~\cite{kafle2019challenges, zhou2020unified, yu2021vision}. One fundamental task, image captioning, which involves the model to generate an informative textual caption according to a given image, opens up a venue for creativity to be explored. Humans can compose concise descriptions of pictures by focusing on what they find important.
~\citet{mao2014deep, xia2021xgpt, mokady2021clipcap, radford2021learning} lay a solid foundation on the current capability of machine learning models to relay cross-modal knowledge for the language models to do generation out of images. 
Beyond generating captions, creating stories--which utilize linguistics to compose and narrate an interrelated series of events~\cite{li-etal-2018-generating, peng2018towards, chandu2019my}--according to a single input image offers even possibilities for creativity-based tasks~\cite{wang2020storytelling, yang2019knowledgeable, hsu2018using}.

\begin{figure}[t]
\centering
\fbox{\includegraphics[width=1.0\linewidth]{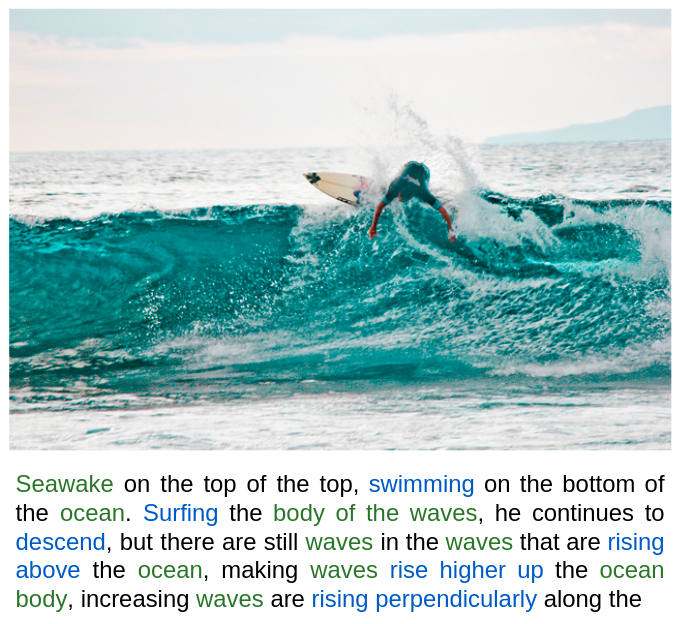}}
\caption{A story generated by action-styled PPST. \textcolor{green}{\textbf{Green}} denotes the words associated with the image. \textcolor{blue}{\textbf{Blue}} denotes the words associated with the style.}
\label{fig:front-sample}
\end{figure}

From the recent advancement on the image-to-story task, it is evident that multiple challenges still remain to be properly solved. One of the main challenges is that model-generated stories tend to lose their coherence as their length increases. Furthermore, the generated text needs to go beyond the pure description of an image as captioning does. Data scarcity, in this context the lack of ready-to-use datasets of image associated with a short story, is also a challenge. Lastly, to the best of our knowledge, there is still limited control over generated stories aside from their relevance to the corresponding image, especially with regards to style~\cite{alabdulkarim2021automatic}. Style has a role to convey a message or story through certain variations of diction and ways of delivery appropriate for a specific context~\cite{ficler-goldberg-2017-controlling,shen2017style,rishes2013generating}.

In this work, we introduce Plug-and-Play Story Teller (PPST). We take a step towards generating a stylistic story from an image while alleviating the data scarcity issue by leveraging large pre-trained models such as CLIP \cite{radford2021learning} and GPT-2 \cite{radford2019language}, and to add the possibility to control the rendered style through plug-and-play adapters, explored in \cite{madotto-etal-2020-plug} and \cite{radford2021learning}. PPST yields improved natural and on-topic stories, and the resulting stylistic stories also have a strong image-story relevance. Our results highlight the performance of PPST, especially in story coherence and image-story relevance, improving the previous state-of-the-art performance. 
Lastly, we provide an analysis on the generated stories, including the occurring issues such as repetition and lack of common sense. We present an example of our generated stories using PPST in Figure \ref{fig:front-sample}.

\section{Related work}

\subsection{Vision-language generation}


In vision-language generation, we exploit both image and text as cross-modal knowledge to address various tasks. Taking on the fact that humans can prepare concise descriptions of pictures by focusing on what they find important,~\citet{mao2014deep} explore this direction by developing a multimodal recurrent neural network model (RNN) to generate novel image captions. ~\citet{xia2021xgpt} build a method of cross-modal generative pre-training for text-to-image caption generators through multiple generation tasks. ~\citet{huang2019attention} build an Attention on Attention (AoA) module, which extends conventional attention mechanisms to determine the relevance between attention results and queries. In the encoder, AoA helps to rectify model relationships among different objects in the image; in the decoder, AoA filters out irrelevant attention results and keeps only the useful ones. 

Further,~\citet{pan2020x} introduce a unified X-Linear attention block, that fully employs bilinear pooling to selectively capitalize on visual information or perform multimodal reasoning to leverage high order intra- and inter-modal interactions.~\citet{cornia2020meshed} build a meshed transformer with memory architecture that improves both the image encoding and the language generation steps. It explores a multi-level representation of the relationships between image regions integrating learned a priori knowledge, and uses a mesh-like connectivity at decoding stage to exploit low- and high-level features.~\citet{mokady2021clipcap} show the effectivity of the encoding from a recent advancement on vision-language pre-training approach, CLIP~\cite{radford2021learning} encoding as a prefix to the caption for image captioning. 

\subsection{Modeling on low-resource data}

Modeling on low-resource data tends to lead to overfitting, which results in non-robust and overly-specific models. This problem is often solved by using augmentation methods. Different augmentation methods and toolkits for various data formats have been developed to better regularize models and increase robustness~\cite{luis2017imageaug,park2019spec,dhole2021nlaug,lovenia-etal-2022-clozer}. 

With the rise of large pre-trained models, astonishing progress has been made for handling low-resource data. Large pre-trained models, such as BERT~\cite{devlin-etal-2019-bert}, GPT2~\cite{radford2019language}, and CLIP~\cite{radford2021learning} have shown to be effective for handling multiple low-resource tasks~\cite{wilie-etal-2020-indonlu,cahyawijaya-etal-2021-indonlg,winata2022nusax,winata-etal-2021-multilingual}. The labelled data for image-to-story task is also scarce, hence we extend these large pre-trained models to allow a more robust image-to-story generation.

\subsection{Image-grounded story generation}

In the image-grounded story generation task~\cite{rameshkumar2020storytelling, wang2020keep, wang2018no, concepcion2016mining, ferraro2019proceedings, mitchell2018proceedings, Min2010image2story}, the widely adopted pipeline includes: 1) extracting captions from an image, 2) encoding the caption, 3) altering the caption with pre-trained encoded stories, and 4) decoding the resulting story. Skip-thought vectors~\cite{kiros2015skip} and a sentence encoder-decoder have been used to build an image-to-story generator or to align books and movies~\cite{Zhu_2015_ICCV}.~\citet{ba2016layer} design alternative pipelines by chaining a convolutional neural network (CNN) to extract feature and a recurrent neural network (RNN) with attention for story generation. 

Other previous works have explored the use of a graph-based architecture \cite{wang2020storytelling} for visual storytelling by modeling the two-level relationships on scene graphs.~\citet{yang2019knowledgeable} present a commonsense-driven generative model, which aims to introduce commonsense from an external knowledge base for visual storytelling.~\citet{hsu2018using} propose an inter-sentence diverse beam search to produce expressive stories. One of the latest works in the field is Image2Story~\cite{Min2010image2story}, which will be further explained in \S\ref{sec:baseline}.

\begin{figure*}[t!]
    \centering
    \fbox{\includegraphics[width=\linewidth, trim={10cm 12.5cm 10cm 12.5cm}, clip]{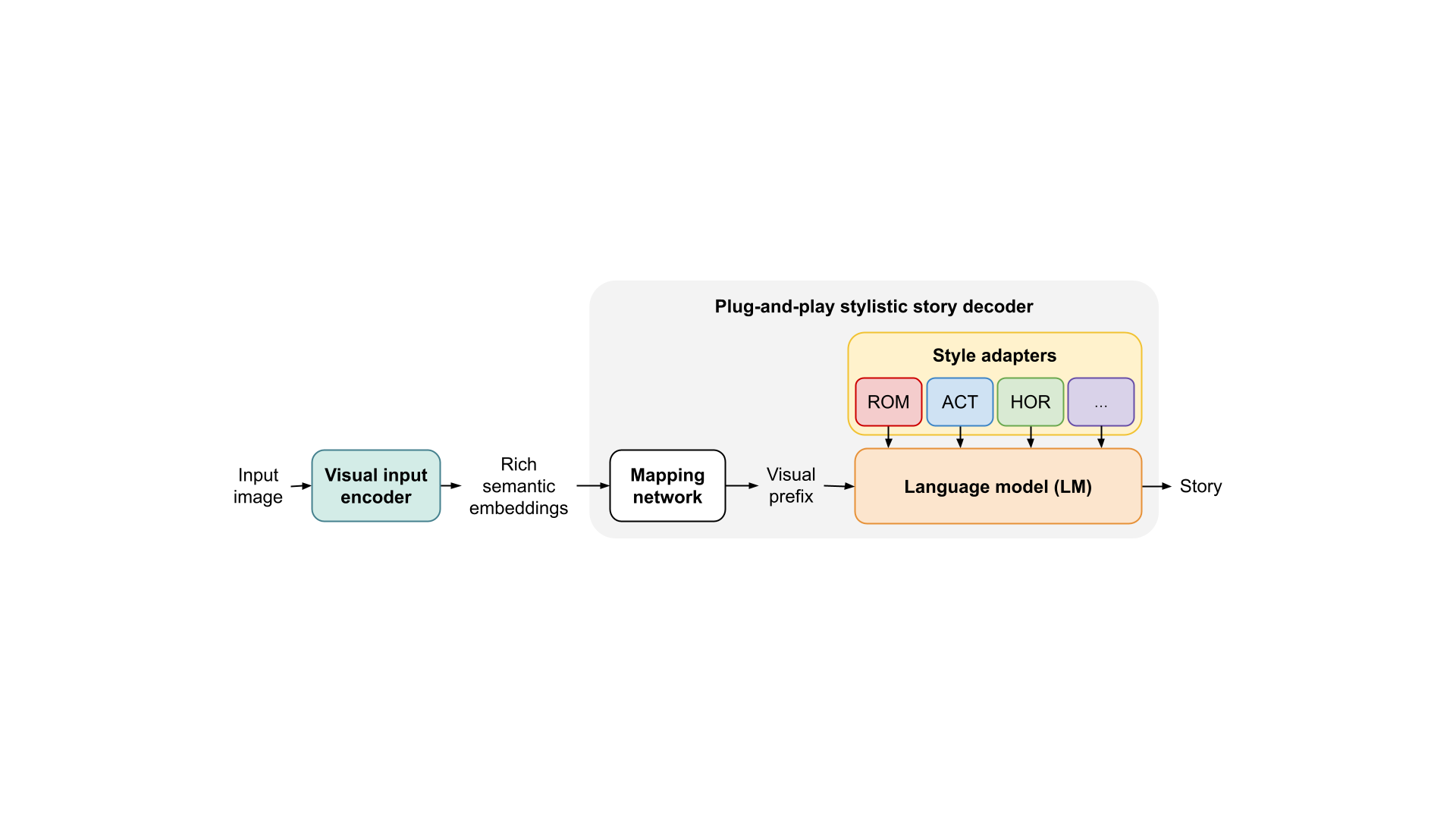}}
    \caption{The inference pipeline of Plug-and-Play Story Teller (PPST) for controllable story generation based on a single image. Visual input encoder encodes the image into a rich semantic embedding, which is then projected into a fixed length visual prefix to be fed to the stylistic language model in order to generate a styled story.
    }
    \label{fig:proposed-system}
\end{figure*}

\subsection{Controllable text generation}

One important aspect required in natural language generation is the control over the produced result. Recent approaches on style generation control have shown promising results.~\citet{dathathri2020Plug} develop plug-and-play language models (PPLM), which combine a pre-trained LM with one or more simple attribute classifiers that guide text generation without any further training of the LM.~\citet{smith2020controlling} adapt \cite{weston2018retrieve, roller-etal-2021-recipes}, and compare it with some of the previously mentioned approach on controlling the styles of generative models to match one among about 200 possible styles.

While~\citet{smith2020controlling} mention that PPLM-style approach is cheaper at train time,~\citet{madotto-etal-2020-plug} highlight its considerable computational overhead.~\citet{madotto-etal-2020-plug} tackle this issue by developing a plug-and-play conversational model (PPCM) that uses residual adapters~\cite{houlsby2019parameter} and discards the need of further computation at decoding time and any fine-tuning of a large LM. At the same time, the generation result using PPCM is also more fluent and style-consistent. For this reason, we adapt PPCM to introduce style controllability into our method.

\section{Plug-and-Play Story Teller (PPST)}
\label{sec:technical-approach}


We present the overview of our approach: Plug-and-Play Story Teller (PPST)
during inference in Figure \ref{fig:proposed-system}. To generate stories out of an image, PPST involves two main components: visual input encoder ($Enc$) and plug-and-play stylistic story decoder ($Dec$). We use two datasets: an image captioning dataset $\mathcal{D} = \{(v^{\mathcal{D}}_i, c^{\mathcal{D}}_i)\}^{n}_{i = 1}$, where $v^{\mathcal{D}}$ denotes image as the visual content and $c^{\mathcal{D}}$ denotes the caption with the textual description of the respective image, and a book passage collection $\mathcal{B} = \{(p^{\mathcal{B}}_i, g^{\mathcal{B}}_i)\}^{n}_{i = 1}$, where $p^{\mathcal{B}}$ denotes the passage chunk and $g^{\mathcal{B}}$ denotes its style (genre).

\subsection{Visual input encoder}
\label{sec:visual-input-encoder}

Initially, PPST needs to be able to grasp what the image depicts on a factual basis (e.g., objects, performed actions, and the implied associations) so it should have prior knowledge to develop the story on. For this purpose, we use CLIP~\cite{radford2021learning}, which learns and accumulates knowledge of visual concepts through a wide variety of image-sentence pairs. CLIP builds its comprehension of text-image alignment by pre-training an image encoder and a text decoder together, and employs a contrastive learning objective to maximize the cosine similarity for the correct image-sentence pairings.
Leveraging the text-image alignment capability provided by CLIP, we utilize its image encoder as the visual input encoder ($Enc$) to produce rich semantic embeddings $\mathcal{R}^\mathcal{E} = \{r^{\mathcal{E}}_i\}^{n}_{i = 1}$ from the images $\{v^{\mathcal{D}}_i\}^{n}_{i = 1}$.

\paragraph{Mapping network}

Although $Enc$ and $Dec$ have been pre-trained using natural language supervision, both of them undergo the learning process separately, which leads to develop latent spaces that provide crucial knowledge but are independent from each other. Furthermore, $Dec$ has yet to be familiar with the visual content offered by the representations generated by $Enc$ ($\mathcal{R}^\mathcal{E}$). To align $Dec$ with the latent space where $\mathcal{R}^\mathcal{E}$ is in, the straightforward way is to simply fine-tune $Dec$ on $\mathcal{R}^\mathcal{E}$.

However, this method expands the number of parameters that $Dec$ has and adds a notable amount of computation cost to the training process. Due to this reason, following~\cite{mokady2021clipcap, li2021prefix}, we introduce a mapping network $Map$ to act as a bridge between the latent spaces of $Enc$ and $Dec$. Using $\mathcal{R}^\mathcal{E}$ as its input, we train $Map$ to produce a fixed length visual prefix $\mathcal{P}^\mathcal{E}$ adjusted to the latent space of $Dec$, so $Dec$ can receive and understand visual information from the prefix $\mathcal{P}^\mathcal{E}$, making fine-tuning on $\mathcal{R}^\mathcal{E}$ more of an option rather than a necessity. The usage of $Map$ in our pipeline is further explained in \S\ref{sec:story-decoder}.

\subsection{Plug-and-play stylistic story decoder}
\label{sec:story-decoder}

Borrowing the natural language 
ability that large pre-trained models possess, we utilize a pre-trained language model $LM$ as a foundation for generating text in our story decoder $Dec$. Utilizing a pre-trained language model lets the generation leverage a large amount of unlabelled texts with a causal language modeling objective.

To equip our story decoder $Dec$ with stylistic capabilities, we follow PPCM~\cite{madotto-etal-2020-plug} approach, by inserting residual adapters \cite{houlsby2019parameter,bapna2019simple} on top of each transformer layer of $LM$. The adapters act as style adapters $StyAdp = \{S_j\}^{m}_{j=1}$ which are responsible for guiding $LM$'s text generation according to the style in use. Each adapter block $S_j$ consists of a layer normalization~\cite{ba2016layer} for efficient adaptation, followed by an auto-encoder~\cite{hinton1993autoencoders} with a residual connection.

\begin{figure}[t]
    \centering
    \fbox{\includegraphics[width=1.0\linewidth]{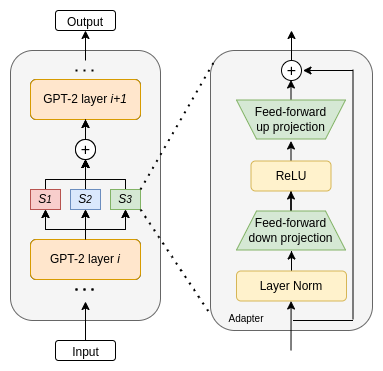}}
    \caption{Architecture of our plug-and-play stylistic language model $SLM$ using GPT-2 language model $LM$ and style adapters $StyAdp$.}
    \label{fig:ppcm}
\end{figure}

For each style from $j = 1$ to $m$, we first select a subset of $\mathcal{B}$ where $g^{\mathcal{B}}_i$ equals the $j$-th style, then train $S_j$ using frozen $LM$ parameters and trainable $S_j$ parameters on the passages in the subset $p^{\mathcal{B}_j}$.
After training, the $StyAdp$ are then utilized to steer the output of the $LM$ distribution at inference time without modifying the original weights. We refer to the $LM$ with the trained $StyAdp$ as plug-and-play stylistic language model ($SLM$). The architecture of $SLM$ is shown in Figure \ref{fig:ppcm}.

Without any modification, an $LM$ 
is conditioned on 
a textual input to prompt text generation. To enable 
$Dec$ to produce texts based on visual representations, we employ $Map$ to translate $\mathcal{R}^\mathcal{E}$ to the input embedding space of $SLM$. During a forward pass, $Map$ projects $\mathcal{R}^\mathcal{E}$ into fixed length visual prefixes $\mathcal{P}^\mathcal{E} = \{p^{\mathcal{E}}_{i}\}^{n}_{i = 1}$ which is then fed to the $Dec$ to perform text generation based on $\mathcal{P}^\mathcal{E}$. By using this pipeline, we train $Map$ using $D$ to allow $Map$ to project meaningful semantic from $\mathcal{R}^\mathcal{E}$ into the input embedding space of $LM$. By combining $Map$ and $StyAdp$, we enable $SLM$ to ground its text generation based on a visual content under a weak supervision introduced by $\mathcal{D}$.

\section{Experiment}
\label{sec:exp}

\subsection{Dataset}
\label{sec:dataset}

As described in \S\ref{sec:technical-approach}, we utilize two types of datasets. The first dataset is related to images and captions. We use MS-COCO~\cite{lin2014microsoft} as our image captioning dataset $\mathcal{D}$. MS-COCO is a large-scale 328K-image dataset commonly used for object detection, segmentation, and captioning. We use the image-caption pairs to obtain prefixes and text embeddings to train the mapping network (see \S\ref{sec:technical-approach}). Due to our computing resource limitation, we utilize only 10\% of MS-COCO total data.

The second dataset is related to books and genres. For the passage collection $\mathcal{B}$, we use BookCorpus~\cite{Zhu_2015_ICCV} to enable the adaptation of generated stories to a prompted genre. BookCorpus is a large dataset composed of 11,038 books adding up to nearly 985 millions words (1.3 millions unique words) used to train large models such as BERT~\cite{devlin-etal-2019-bert}. We obtain the styles of the books by matching the book titles in BookCorpus
with the genres in 2021 Smashwords~\cite{bandy2021addressing}
dataset. Smashwords is a dataset listing the e-books available on the Smashwords platform and recording their title, language, price, publication date, URL, and genre.

As a result, we classify the books in 16 genres: romance, fantasy, science fiction, new adult, young adult, thriller, mystery, vampires, horror, teen, adventure, literature, humor, historical, themes, and other.
Finally, we split the book texts based on paragraphs, select the text chunks that consist of 30-60 words as passages, and discard the rest. The total number of passages in our dataset nears 7.7M.

\subsection{Experiment setup}

We use a pre-trained CLIP with Vision Transformer encoder
to obtain text-image alignment representation as $Enc$. We note that different from the settings used in~\cite{madotto-etal-2020-plug}, where they use open-domain generic dialogues to serve as a prefix to trigger the responses, here we use a visual prefix to trigger the generation in our experiments. Due to the difference in use case, and to enable tendency towards longer generation responses, we use a GPT-2 model instead of the proposed utilisation of DialoGPT~\cite{zhang2019dialogpt} in~\cite{madotto-etal-2020-plug}. In detail, as for the $LM$ in \S\ref{sec:story-decoder}, we utilize a pre-trained GPT-2 with 124M parameters,
and employ the same model architecture and size as well for the adaptation of~\cite{madotto-etal-2020-plug}.

We conduct the experiment using PPST with a non-stylistic setting (without style adapter), referred to as \textbf{Non-styled}, and with two stylistic settings, which are \textbf{Romance} and \textbf{Action}, since they are the styles represented by most amount of samples in the BookCorpus dataset. To filter out the samples that is strongly categorized as \textbf{Romance} and \textbf{Action}, we use the first three genres listed by BookCorpus entries to recognize those entries as \textbf{Romance} and \textbf{Action} entries.

For \textbf{Non-styled}, we utilize the same approach described in \S\ref{sec:technical-approach}, but instead of using an LM guided by a style adapter, we use a regular pre-trained LM (no style adapter) $LM$ directly fine-tuned on the book collection. We employ \textbf{Non-styled} as a comparison against the stylistic approaches in terms of a controllable story generation. For \textbf{Romance} and \textbf{Action}, fine-tuning of the GPT-2 with style adapters on the book collection data is done for a maximum of 10 epochs, with a learning rate of 1e-3, a batch size of 8, and a maximum sequence length of 512. During the training on image-sentence pairs, we only train the mapping network with a prefix size of 512, a prefix length of 10, and an activation function of \texttt{tanh}, and freeze the LM.

Our story generation employs beam search with a beam size of 5, a temperature of 0.8, and a top-k of 10. To avoid repetition, we apply a repetition penalty of 0.7 and limit any repetition of 3-gram phrases. To encourage the model to produce a longer story, we apply an exponentially decaying length penalty with a factor of 1.7 after 20 tokens and set a minimum generation length to be 750.


\subsection{Baseline}
\label{sec:baseline}



We use \textbf{Image2Story}~\cite{Min2010image2story} as our baseline. It combines an RNN and encoder-decoder structure to generate a short story out of an image. The model is built upon skip-thought encoders and structured in a 3-stage pipeline where: 1) a caption based on an input image and a skip-thought vector based on an image-caption dataset are created, 2) a skip-thought vector based on a story dataset is created, and 3) starting from the caption, the vector in 1) is subtracted and the vector in 2) is added so as to obtain a story fitted to the story dataset based on the input image.

\subsection{Evaluation setup}
\label{sec:eval}


PPST relies on visual semantics and information, so we need to ensure that they manage to extract sufficient knowledge from the input image.
For this purpose, we use the original captions provided from the MS-COCO dataset as gold references representing the visual content conveyed by the input images for the text-to-text similarity metrics, and the images for the image-to-text similarity metric.

\paragraph{Automatic evaluation} We compute seven automatic evaluation metrics covering two n-gram-based text-to-text similarity metrics, i.e,  ROUGE-L~\cite{lin-2004-rouge} and  ChrF++~\cite{popovic2017chrf}; four model-based text-to-text similarity metrics, i.e., MoverScore~\cite{zhao2019moverscore}, BERTScore~\cite{zhang2020bertscore}, BLEURT~\cite{sellam2020bleurt}, and BARTScore~\cite{yuan2021bartscore}; and one image-to-text  similarity metric, i.e. CLIPScore~\cite{hessel2021clipscore}. For the image-to-text similarity, we compare the text directly with the original image used for generating the story.

\paragraph{Human evaluation} To further assess the quality of the generated stories from our system, we conduct a human evaluation in addition to computing the metrics previously mentioned. Each participant is given a questionnaire composed of 10 subsections. Each subsection has 1 image, randomly sampled from our dataset, followed by four stories respectively generated by 1) \textbf{Image2Story}, 2) our \textbf{Non-Styled} model, 3) our \textbf{Romance} model, and 4) our \textbf{Action} model. For all models, we ask if "the story makes sense" to assess story coherence, and if "there is a link between the image and the story" to assess image-story relevance. In addition, for our \textbf{Romance} and \textbf{Action} models, we ask a third question to know if "the story has the given style" to judge style fitness. The participants answer to the questions using a 5-point Likert scale with the choices: "A lot", "A little", "Neutral", "Not really", and "Not at all".
The human evaluation is conducted on 13 participants.


\section{Result and analysis}
\label{sec:result-and-analysis}

\begin{table*}[t]
    \centering
    \resizebox{1.0\linewidth}{!}{
        \begin{tabular}{lccccccc}
        \toprule
        \textbf{Model} & \textbf{ROUGE-L} & \textbf{ChrF++} & \textbf{MoverScore} & \textbf{BERTScore} & \textbf{BLEURT} & \textbf{BARTScore} & \textbf{CLIPScore} \\
        \toprule
        \textbf{Image2Story} & 9.06 & 15.04 & 50.20 & 39.65 & 23.80 & -4.00 & 59.95 \\
        \midrule
        \textbf{PPST Non-styled} & 9.52 & \textbf{16.81} & 50.11 & 39.71 & 26.51 & -4.05 & 61.99 \\
        \textbf{PPST Romance} & 10.02 & 15.30 & \textbf{51.94} & 46.48 & \textbf{36.70} & \textbf{-3.86} & 69.02 \\
        \textbf{PPST Action} & \textbf{10.09} & 15.28 & \textbf{51.94} & \textbf{46.59} & 36.69 & \textbf{-3.86} & \textbf{69.21}\\
        \bottomrule
        \end{tabular}
    }
    \caption{Automatic evaluation results on the visual information retention in the generated stories. For image-to-text similarity, i.e., CLIPScore, we compare the generated stories directly with the corresponding images, while for text-to-text similarity metrics we use the original captions provided from the MS-COCO dataset.}
    \label{tab:preliminary-results}
\end{table*}

\begin{figure*}[t]
\centering
\fbox{\includegraphics[width=0.31\linewidth, scale=0.5, trim={0.5cm 0.43cm 0.5cm 0.43cm}, clip]{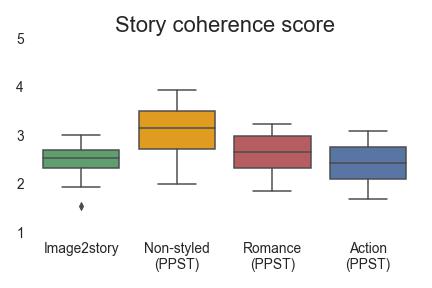}}
\fbox{\includegraphics[width=0.31\linewidth, trim={0.5cm 0.43cm 0.5cm 0.43cm}, clip]{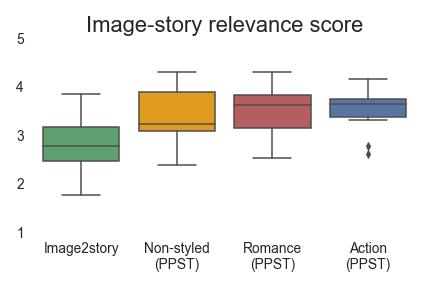}}
\fbox{\includegraphics[width=0.31\linewidth, trim={0.5cm 0.43cm 0.5cm 0.43cm}, clip]{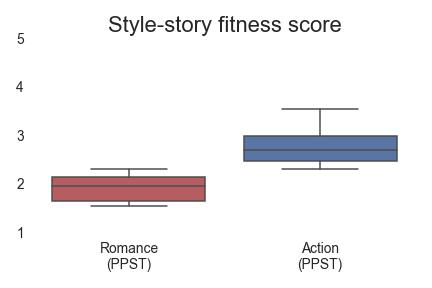}}
\caption{Human evaluations of the generated stories from all models in terms of story coherence \textbf{(left)},  image-story relevance \textbf{(middle)}, and  style-story fitness \textbf{(right)}.}
\label{fig:human_evaluation_likert.jpg}
\end{figure*}



\subsection{Image-to-story generation quality}

As explained in \S\ref{sec:eval}, we utilize both automatic and human evaluation to measure the quality of the generated story of four models: 1)~\citet{Min2010image2story}'s \textbf{Image2Story}, 2) our \textbf{Non-styled}, 3) our \textbf{Romance}, and 4) our \textbf{Action}. Table~\ref{tab:preliminary-results} shows all the automatic evaluation metrics of the generated story. In general, all of our models outperform the baseline \textbf{Image2Story} in both n-gram-based text-to-text similarity, model-based text-to-text semantic similarity, and image-to-text semantic similarity metrics. More specifically, the \textbf{Romance} and \textbf{Action} models perform significantly better on semantic text-to-text and image-to-text similarity metrics by $\sim$7\% on the BERTScore, $\sim$10\% on the BLEURT, and $\sim$8\% on the CLIPScore. The \textbf{Non-styled} model performs not as good as the \textbf{Romance} and \textbf{Action} models but still yields a slightly better score compared to the \textbf{Image2Story} model in most metrics. This automatic evaluation result suggests that PPST, with and without the style adapter, can generate a better image-grounded story despite having no direct supervision for the image-to-story generation task itself.

 

The human evaluation result is shown in Figure \ref{fig:human_evaluation_likert.jpg}. In terms of coherence, our evaluation result suggests that stories generated by \textbf{Non-styled} surpasses all other models, with an average rating of 3.12, followed by \textbf{Romance}, \textbf{Image2Story}, and \textbf{Action}). This suggests that pre-trained LM is sufficient to generate coherent stories without requiring tuning on the sentence-to-story generation task as incorporated in the prior work~\cite{Min2010image2story}, which shows PPST performs well despite the image-story data scarcity issue. 

The relevance between the image and the story aligns with the automatic evaluation result. Our models, especially the stylistic \textbf{Romance} and \textbf{Action}, outperform the baseline \textbf{Image2Story} by a large margin, achieving a rating score $3.5$ compared to only 2.77, which suggests a better text-image alignment compared to the prior work.
For style fitness, we find that our \textbf{Action} model achieves an adequate style-story score of 2.78, while the \textbf{Romance} model, only obtain a romance style-story score of 1.91. We further explicate this phenomenon in \S\ref{sec:analysis}.


\subsection{Analysis on the generated stories}
\label{sec:analysis}

Aside from the automatic and human evaluations, we manually inspect the stories to gather insights regarding the behaviors of our models. Table \ref{tab:story-samples} provides 2 examples of our image-to-story generation. Similar to the majority of the book passages used in the training step, our generated stories are inclined to lean towards \textit{describing} the visual aspects of the input image and slowly building the occurring events from there, which notably accounts for PPST's higher image-story relevance scores, rather than recounting a chain of events or actions in a straightforward manner as the baseline. We also find that while the surrounding contexts of our stories are relevant to the respective images, this relevance deteriorates as the stories grow longer.

Furthermore, we observe that our styled generation result can contain repetitions and tends to use a few words more often than the others.
This aligns with the drawbacks of PPCM described by~\citet{madotto-etal-2020-plug}, which are mainly caused by the restricted use of vocabulary for generating attribute consistent responses. It is also mentioned that this abuse of restricted vocabulary harms fluency, because it cannot always fit within a given context. 
All these limitations negatively impact the coherence and fluency, therefore the overall quality of the generated stories.
Finally, in spite of the proven capability of controlling the generation, the style-story score, on the right plot of Figure \ref{fig:human_evaluation_likert.jpg}, shows that there is still potential for improvement. We leave this exploration for future work, specifically on realizing more generation control, in this case by improving the generated stories to be more related to the styles being adapted.

\begin{table*}[t!]
    \centering
    \resizebox{\linewidth}{!}{%
    \begin{tabular}{p{0.2\linewidth} p{0.8\linewidth}}
        \toprule
            \textbf{Image} & \textbf{Generated stories} \\
        \toprule
            \multirow{4}{*}{\raisebox{\totalheight}{\includegraphics[width=\linewidth]{images/000000112798.jpg}}} & \textbf{Image2Story:} a a cat on a laptop top . Alice lay down , trying to decipher . She was a cat , too , so he did n't even know what to do . She resisted , yet she found herself wanting to keep her head on top of a computer . With every step she laid on top of a cat that had taken on as long as you could , I snapped my laptop shut . I lay on my computer , a reminder of a cat that had a job . \\
             & \textbf{PPST Non-styled:} A cute cat reading a computer and a pad on a desk in a office. 'This is the best I've seen in a long while.'  "I don't know how to say this, but I'm not going to let you do this to me. You're the one that's going to have to deal with me. I don \\
             & \textbf{PPST Romance:} a cats pawing a tree is lying on a table. ``I don't have a computer or computer monitor. I do have some photographs taken during the day. I am trying to find some time to spend together. I have been working on some photographs that show how long the cat has been sitting in the same room as the computer. Next to \\
             & \textbf{PPST Action:} there is a cat on a table. and next to the table, a moving hand is used a hand. and the monitor is open. The cat is lying on the table with its legs folded up. The monitor is tilted slightly upward toward the table itself. The table itself is folded up slightly slightly from the table's edge to the floor surface. \\
            \midrule
            \multirow{4}{*}{\raisebox{\totalheight}{\includegraphics[width=1.0\linewidth]{images/000000343803.jpg}}} & \textbf{Image2Story:} a a man is nothing . I was wrestling a picnic , he shouted . I had known he was going to throw his leg around my ankle , and Nate gave it to me . It s a kid , as a friend of mine . In the beginning , I realized I had been carrying her to the field and carrying him on a leash , making it all feel like a baseball game . I would practically twitch before he was responsible . \\
             & \textbf{PPST Non-styled:} A human being that is having fun!  It was the first time I've seen someone that I really like.  I hope I'm doing the right thing.  It wasn't long ago I was a stranger that someone I lost my virginity to."Well, I'm not sure, but I'm sure it's not the same. I \\
             & \textbf{PPST Romance:} A person who is in the Frisbee field with two Frisbees in hand. In contrast to the frisbee riguring to catch a Frisbier. Behind them, however, there appears to be nothing unusual happening. Rather, it appears that there is little fuss that happens this year. However, nothing unusual has transpired this year \\
             & \textbf{PPST Action:} A person who is in the Frisbee field with two frisbees. In the air they are holding them. Behind them stands a young man holding a frisb. They are holding Frisbees in their legs. Hands are placed over their necks to allow them to sit comfortably. They sit comfortably in their chairs to sit upright. They \\
        \bottomrule
    \end{tabular}
    }
    \vspace{2mm}
    \caption{Samples of image-to-story generation result generated from \textbf{PPST Non-styled}, \textbf{PPST Romance}, and \textbf{PPST Action} against the baseline \textbf{Image2Story}.}
    \label{tab:story-samples}
\end{table*}

\section{Discussions}

Our works have moved image-grounded story generation forward by improving the generated story coherence and image-story relevance, and by adding a layer of style control on top of it. However, as explained in \S\ref{sec:result-and-analysis}, the current progress still leaves room for improvement. 

\paragraph{Story coherence} Taking inspiration from recent works, a few strategies to refine story coherence can be implemented as the next step, for instance an unsupervised hierarchical story infilling~\cite{ippolito-etal-2019-unsupervised}, a semantic dependency skeleton generation to extract key information~\cite{xu-etal-2018-skeleton} or storyline~\cite{yao2019plan}, a deeper understanding of causal and temporal relations of events through commonsense knowledge~\cite{mostafazadeh2016caters}, the utilization of both sentence-level and discourse-level prefix information for decoding~\cite{guan2020knowledge}, and making use of a story dataset with rich and fine-grained annotations~\cite{akoury2020storium}.

\paragraph{Image-story relevance} For the relevance, rather than simple embedding concatenation, other ways to incorporate visual information to textual~\cite{liu2019aligning} and deepen visual comprehension~\cite{fang2015captions, huang2016visual} can be further investigated.

\paragraph{Style control} We also highlight the interesting directions in advancing the realization of control over stylistic story generation. Our exploration underlines the importance of improving generated stories to relate more to the styles being adapted. Improved and new approaches to control the generated stories with more specific, descriptive, even depicted by a short passage, styles will open up interesting venues on controllable text generations to assist artistic and creative tasks, whether these methods include a pre-trained model \cite{keskar2019ctrl, gan2017stylenet, hu2017toward} or not \cite{hu2020text}.

\section{Limitations}


We discuss here about the limitations of our work, specifically concerning the chosen heuristic to align style and passages from BookCorpus, the limited amount of data in choice of style for the adapters, and possible biases. 

As explained in section \ref{sec:dataset}, we choose to split book texts based on paragraphs in order to retain a certain degree of logical fluency throughout each story samples, as a paragraph usually deals with a single theme or idea. While this helps keeping the passages relatively short, one limitation of this approach is that some passages might not fully reflect the style of narrative that it is classified as. For example, not every paragraph taken out of context from a romance book will exhibit its genre. There could be a sizable amount of passages that focus on world-building and laying the groundwork for the book's main plot to progress.

We decide to focus on romance and action because these styles are the most represented in the dataset used, as well as being more straightforward to capture in terms of style compared to other genres that rely on an underlying plot throughout the book such as historical or adventure. Generalizing PPST to these styles with a lower amount of resources might require further experiments.

Lastly, previous works have shown that captioning models can exhibit harmful biases, such as gender bias~\cite{hendricks2018women} and racial bias~\cite{zhao2021understanding}. Since we pair those image captions data with written stories from a wide variety of books, those biases can be further amplified. Thus, such generative processes must be used with caution. While tackling unwanted biases in images or captions is a must, the bias exhibited in stories is sometimes justified by the context and the surrounding narrative. Not all stories should be completely neutral, and this balance should be considered carefully in future directions.

\section{Conclusion}

By leveraging text-image alignment representations to describe the visual content of a given image in words, we can use the resulting semantic embeddings as prior knowledge to generate a short story out of a given picture through a plug-and-play controllable language model approach. It also allows us to tackle the data scarcity issue in this task. 

The results show that our Plug-and-Play Story Teller (PPST) generates more consistent and on-topic stories according to the visual information, as well as performing better in relevance and image-story relationship than the previous state-of-the-art. We also found that PPST without style adapters (\textbf{Non-styled}) generates more coherent stories, and PPST utilizing style adapters (\textbf{Romance} and \textbf{Action}) have a similar, if not a slightly better, image-story relationship than the other approaches.




\bibliography{anthology,custom}
\bibliographystyle{acl_natbib}




\end{document}